\pdfoutput=1

\documentclass[11pt]{article}

\usepackage[preprint]{coling}

\usepackage{times}
\usepackage{latexsym}

\usepackage[T1]{fontenc}

\usepackage[utf8]{inputenc}

\usepackage{microtype}

\usepackage{inconsolata}

\usepackage{graphicx}

\usepackage{tempora}

\usepackage[english, russian]{babel}
\usepackage[T1, T2A]{fontenc}
\usepackage{CJKutf8}
\usepackage[utf8]{inputenc}
\usepackage{caption}
\usepackage{subcaption}
\usepackage{multicol}
\usepackage{placeins}
\usepackage{titlesec}
\usepackage{float}

\usepackage{enumitem}
\usepackage{tabularx}

\usepackage{hyperref}
\usepackage[framemethod=tikz]{mdframed}
\usepackage{booktabs}

\AtBeginDocument{\renewcommand{\abstractname}{Abstract}}

\title{CleanComedy: Creating Friendly Humor through Generative Techniques}

\author{
 \textbf{Dmitry Vikhorev\textsuperscript{1}},
 \textbf{Daria Galimzianova\textsuperscript{1,2}},
 \textbf{Svetlana Gorovaia\textsuperscript{1}},
\\
 \textbf{Elizaveta Zhemchuzhina\textsuperscript{1}},
 \textbf{Ivan P. Yamshchikov\textsuperscript{3}}
\\
 \textsuperscript{1}LEYA Lab,  HSE University,
 \textsuperscript{2}MTS AI,
 \textsuperscript{3}CAIRO, THWS
}

\begin{document}
\maketitle
\begin{abstract}
Humor generation is a challenging task in natural language processing due to limited resources and the quality of existing datasets. Available humor language resources often suffer from toxicity and duplication, limiting their effectiveness for training robust models. This paper proposes CleanComedy, a specialized, partially annotated toxicity-filtered corpus of English and Russian jokes collected from various sources. We study the effectiveness of our data filtering approach through a survey on humor and toxicity levels in various joke groups. In addition, we study advances in computer humor generation by comparing jokes written by humans with various groups of generative jokes, including our baseline models trained on the CleanComedy datasets.

\end{abstract}

\section{Introduction}

Computational humor is a significant area of research within natural language processing. The humor analysis is complex due to its reliance on contextual dependencies. Despite this challenge, computational humor offers possibilities for improving human-computer interaction. 

Previous works in this field \cite{chen-soo-2018-humor, yang-etal-2015-humor, mihalcea-strapparava-2005-making} have laid the baseline in humor recognition. Recent research has focused on developing humor generation solutions based on LLMs \cite{jentzsch-kersting-2023-chatgpt, amin-burghardt-2020-survey}. 

In this study, we collect, filter and prepare humor datasets in English and Russian languages. In order to further enhance the utility of the datasets, we collect human humor scores for 1,000 Russian and 1,000 English jokes. We publish individual annotations for every joke without aggregating scores provided by multiple annotators.

The datasets are publicly available on GitHub\footnote{\url{https://github.com/gorovuha/CleanComedy}}, facilitating access for the research community. 

Following, we fine-tune and align an LLM on the prepared data, generate new humor samples and ask human annotators to evaluate them. The model undergoes two training stages: \textbf{Supervised Fine-Tuning} and \textbf{Alignment} to learn the humor style and subsequently pick up what ``funniness'' is. We use the scores obtained from the human annotators during the alignment stage.

The human judgments are then obtained for results of the two-stage fine-tuning presented in this research. They show that this training technique might give rise to lightweight humorous models and encourage responsible and effective generative AI. 
Although one could cherry pick some fun and interesting jokes, generative humor generally remains an open research problem. 

This paper presents a detailed account of our methodology for dataset collection, filtering, and annotation, as well as provides the settings and results of the experiments with LLM fine-tuned on our dataset in comparison with existing models.

\section{Related work}

Computational humor research has seen significant advancements in recent years.
One of the primary challenges in this field is capturing the details and contextual dependencies that make humor effective and enjoyable.

Works in computational humor \cite{mihalcea-strapparava-2005-making, yang-etal-2015-humor, chen-soo-2018-humor}, lay the groundwork by exploring the field of humor recognition. They introduce linguistic features, which are essential in distinguishing humorous content from non-humorous text. 

Humor often involves a delicate balance of positive and negative emotions, the work \cite{meaney-etal-2021-semeval} explores the dual nature of humor and offense, highlighting the importance of sentiment analysis in humor research.

\subsection{Data}
\label{subsec:data}

The evolution of computational humor research has produced a variety of popular datasets, which serve as the foundational data sources for training humor recognition and generation models. Table ~\ref{Comparison-of-Humour-Datasets} provides an overview of the most commonly referenced datasets. \textbf{16k One-Liners} \cite{mihalcea-strapparava-2005-making, weller2019humor, chen-soo-2018-humor} consists of 16,000 English one-liner jokes sourced from various online collections and 16,000 non-humorous sentences from news titles, proverbs, British National Corpus, and Open Mind Common Sense collection. \textbf{Pun of the Day} \cite{yang-etal-2015-humor} is a collection of puns gathered from the popular website "Pun of the Day." Each entry is user-submitted and typically reviewed by website moderators before inclusion.
\textbf{Short Jokes} \footnote{\url{https://github.com/amoudgl/short-jokes-dataset}} consists of 231,657 brief jokes extracted from various joke websites and social media platforms. \textbf{SemEval 2021 Task 7: HaHackathon, Detecting and Rating Humor and Offense} \cite{meaney-etal-2021-semeval} contains 4,932 jokes, mostly from \textit{twitter}. \textbf{Knowledge Amalgam: Generating Jokes and Quotes Together} \cite{chippada2018knowledge} introduces 96,910 jokes, which was sourced from CrowdTruth and Subreddits, and then deduplicated. It also includes quotes and tweets from different sources. \textbf{The Naughtyformer: A Transformer Understands Offensive Humor} \cite{tang2022naughtyformer} introduces 92,153 jokes divided into three different categories: Clean Jokes, Dark Jokes, and Dirty Jokes. It also includes news texts as negative samples. \textbf{Humor Detection: A Transformer Gets the Last Laugh} \cite{weller2019humor} contains 14,946 jokes scrapped from Reddit. It also includes the \textit{Short Jokes} and \textit{Pun of the Day} datasets listed above.

For the Russian language, there are fewer public datasets with jokes \cite{bolotova-2017-ir}. The most suitable is \textbf{FUN dataset} \cite{blinov-etal-2019-large}, collected from Russian social media and humor resources. It contains different types of jokes, such as one-liners, multi-turn jokes, and short sketches. There is a manual binary annotation of about 2,000 instances. 

These datasets are broadly used in solving computational humor tasks yet they share two major problems that hinder their usage for training of generative humor models:
\begin{enumerate}[nosep]
    \item \textbf{Toxicity}: Many jokes in these datasets contain offensive content, including sexist, racist, or otherwise discriminatory jokes. This does not only raise ethical concerns but also impacts the generalisation capacity of the humor models trained on such data.
    \item \textbf{Redundancy}: There is a high number of duplicate jokes, both explicit when the same exact words are used, and implicit when the same joke is paraphrased multiple times. Training models on duplicate data can lead to overfitting.
\end{enumerate}

\subsection{LLMs}
There has been several recent attempts to explore humor abilities of LLMs.

The study by the authors of \cite{jentzsch-kersting-2023-chatgpt} explores the abilities of ChatGPT 3.5 in both creating and comprehending humor. Their findings indicate that the model struggles with the generation and interpretation of jokes. For instance, ChatGPT 3.5 produces only a limited set of unique jokes, which appear to be hard-coded.

The paper~\cite{mirowski2024robot} studies the LLM as creative writing assistants for comedians and conclude that the models can be deployed successfully in this quality with certain restrictions, such as contextual alignment.

\citet{sunkara2024adaptive} introduced an adaptive humor generation model that personalizes content based on user feedback, demonstrating improved engagement through tailored humor.

\citet{zhang2024improving} emphasized the importance of multimodal context in humor generation, integrating visual and textual data to capture nuanced humor elements.

\citet{li2024contrastive} proposed a novel approach using contrastive learning with humor-contextualized embeddings, resulting in more contextually appropriate and diverse humorous text generation.

\citet{dsilva2024humor} explored augmenting LLMs with humor theories to understand puns, highlighting the challenges LLMs face in grasping linguistic nuances and the potential of theoretical frameworks to improve comprehension.

\citet{wu2024cultural} conducted a comparative study on the role of cultural context in humor generation, revealing significant differences in humor perception between Western and Eastern cultures and underscoring the necessity for culturally aware humor models.

\section{Dataset Curation}

In this section, we describe the methodology employed in collecting and processing our dataset, which includes humor content in both English and Russian languages. The approach to dataset curation is guided by the goal to create a resource that is not only large and varied but also devoid of offensive content and redundancies that could potentially bias or undermine the performance of language models trained on it.

\subsection{Data Processing}
\label{subsec:data_processing}

\begin{table*}[h!]
  \centering
  \begin{tabular}{p{4.5cm}p{6.5cm}ll}
    \toprule
    Name     & Description     & Jokes   & Non-jokes \\
    \midrule
    \multicolumn{4}{c}{English}                   \\
    \midrule
    16k One-Liners  & Humorous samples from daily joke websites    & 16,000 & 16,000 \\
    Short Jokes & Short jokes scraped from various joke websites with lengths ranging from 10 to 200 characters &  231,657  & 0      \\
    SemEval 2021 Task 7: HaHackathon, Detecting and Rating Humor and Offense & 80\% jokes from twitter and 20\% jokes from Short Jokes dataset & 4,932 & 3,068 \\
    Knowledge Amalgam: Generating Jokes and Quotes Together & Multiple sources were combined and deduplicated (the two sources for jokes are CrowdTruth and Subreddits) & 96,910 & 173,633 \\
    The Naughtyformer: A Transformer Understands Offensive Humor & Scrapped from Reddit and includes different types of humour & 92,153 & 10,710 \\
    Humor Detection: A Transformer Gets the Last Laugh & The original part is scraped from Reddit & 14,946 & 0 \\
    Pun of the Day  & Puns from Pun of the Day website  & 2,423 & 2,423 \\
    CleanComedy English    & Ethical filtered jokes with 2-scale score   & 44,481    & 0 \\
    CleanComedy English Gold   & Ethical filtered jokes with human humor 5-scale score  & 1,000 & 0 \\
    \bottomrule
    \multicolumn{4}{c}{Russian}                   \\
    \midrule
    Stierlitz   & One-liners from social media & 46,608 & 46,608    \\
    FUN & Russian jokes from social media and online collections    & 156,605   & 156,605   \\
    CleanComedy Russian    & Ethical filtered jokes with 2-scale score   & 40,926    & 0 \\
    CleanComedy Russian Gold   & Ethical filtered jokes with human humour 5-scale score  & 1,000 & 0 \\
    \bottomrule
  \end{tabular}
    \caption{Comparison of Humor Datasets}
    \label{Comparison-of-Humour-Datasets}
\end{table*}

Firstly, we collect all the data mentioned in Section \ref{subsec:data}, see Table \ref{Comparison-of-Humour-Datasets} for a quick overview. Then we united it into one large collection and removed exact duplicates, ignoring punctuation and case. Every step of the further filtering process removes bad samples. We do not paraphrase the jokes in order to preserve the intricate semantics of humor.

The initial preprocessing of English data involves the removal of examples that contain all symbols except Latin characters and punctuation marks. It helps get rid of bad examples comprising emojis, links and also, for example, corrupted jokes from Reddit containing \textit{[deleted]} or \textit{[removed]} words (these words indicate that a part of the joke was deleted before it was collected).
We also noticed that excessively short or long entries have a large amount of noise (for example, repetitions, unfunny utterances), so the next step was to keep examples only between 50 and 150 characters, see Table ~\ref{en-rubbish} in Appendix. 

For both English and Russian, we utilised unbiased toxicity classifier Detoxify \cite{Detoxify} for extracting toxic jokes as this model is capable of detecting different types of toxicity like threats, obscenity, insults, and identity-based hate. As the authors of the classifier mention, words associated with swearing, insults or profanity are  likely to be classified as toxic, regardless of the tone or the intent. According to our goal, which is to collect an ethical humor dataset, we removed all the jokes that were supposed to be toxic. Due to lower representation of Russian in multilingual models, we employ ruBERTConv Toxic Classifier \footnote{\url{https://huggingface.co/IlyaGusev/rubertconv_toxic_clf}}, a transformer model tailored for the Russian language \cite{kuratov2019adaptation, dementieva2021methods}, to improve the accuracy of toxicity recognition . 

To identify and remove duplicate jokes, we utilize Sentence-BERT (SBERT) pretrained model \textbf{all-mpnet-base-v2} \cite{reimers-2019-sentence-bert, reimers-2020-multilingual-sentence-bert}. This framework allows us to compute text embeddings and then compare their cosine similarity and find sentences with similar semantics \cite{Farouk_2019}. The goal is to remove jokes with the same meaning expressed in different words or jokes where some parts are presented in a different order (see Table ~\ref{duplicates} in the Appendix). As we collect English jokes from various sources, there is a large number of duplicate instances. We consider two entries as duplicates if the cosine between their embeddings is higher than 0.7 for English and higher than 0.9 for Russian. 

After that, rigorous manual analysis of the dataset shows that a large part of jokes contains metaphorical insults. To make the resulting data cleaner we label jokes with zero-shot classifier \textbf{DeBERTa-v3-large-mnli-fever-anli-ling-wanli} \cite{laurer2024less}. The set of labels this time is \textit{politics, neutral, offending, alcohol, drugs, racist}. Thus, we removed all instances labeled as political, racist, and insulting content (see Table ~\ref{deberta} in the Appendix) to avoid unethical expressions.

A thorough analysis of the remaining jokes reveals a significant number of texts on sensitive topics. Therefore, we utilize topic modeling to understand our large corpus better. Our objective is to have topics that are specific enough (e.g., separating soft drinks from alcohol) but not overly fragmented (e.g., having 100 classes about animals would be uninteresting). \textbf{BERTopic}, a topic modeling technique that leverages transformers and c-TF-IDF \cite{grootendorst2022bertopic}, is used to create clusters with easily interpretable topics while keeping important words in the topic descriptions. The visual representation of these clusters can be found in Appendix \ref{fig:clusters-en}, \ref{fig:clusters-ru}. 

Cluster modeling allows us to thoroughly analyze the collected dataset in terms of diversity and content distribution. We remove sensitive jokes that belong to the clusters about \textit{religion, funerals, bathrooms, officers, pregnancy, nations, disabilities, and divorce} ensuring ethical and useful dataset. 

\begin{figure*}[t]
    \centering
    \includegraphics[width=\textwidth]{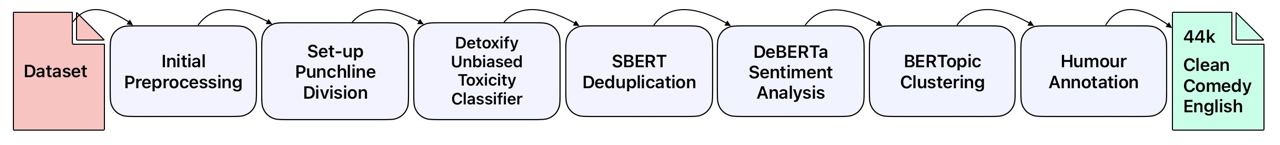}
    \caption{Topic modelling for CleanComedy English.}
    \label{pipe-eng}
\end{figure*}

\begin{figure*}[t]        
\centering
        \includegraphics[width=\textwidth]{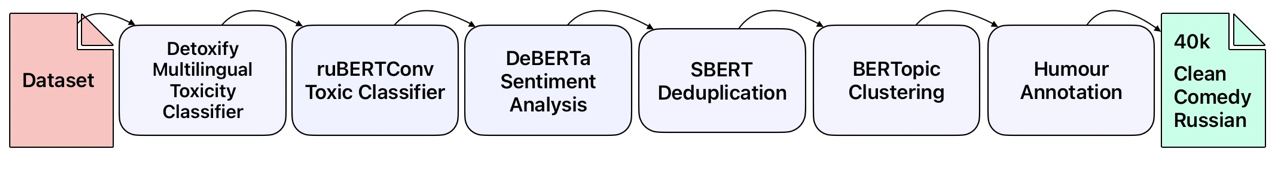}
        \caption{Topic modelling for CleanComedy Russian.}
        \label{pipe-ru}
\end{figure*}

The final remaining dataset, after the full set of filters has been applied, contains 44k English jokes and 40k Russian jokes. The comprehensive filtering pipeline can be found in Figures~\ref{pipe-eng} and \ref{pipe-ru}.

\subsection{Dataset Annotation}
\label{subsec:dataset_annotation}

Datasets for both languages include manual human annotation of 1,000 samples for each language. We call this subset of the data \textbf{CleanComedy Gold}. Volunteers have been asked to rate a joke on a scale from 1 to 5 depending on how funny they believe the joke is. The instruction for the annotation included the following points:

\begin{itemize}[nosep]
    \item Rate how funny a joke is on a five-point scale, where 1 is not funny and 5 is very funny.
    \item Do you find the text of the joke vulnerable or inappropriate? 
\end{itemize}

\begin{figure*}[ht]
    \centering
    \includegraphics[width=0.5\textwidth]{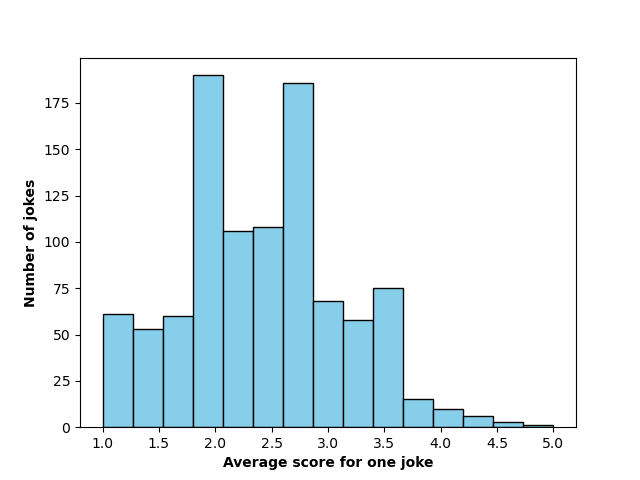}\hfill
    \includegraphics[width=0.5\textwidth]{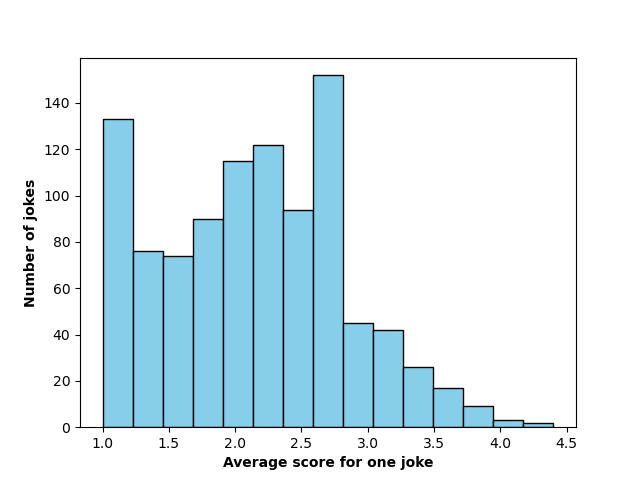}\hfill
    \caption{Average scores for \textbf{CleanComedy Gold} datasets in English (in the left picture) and Russian (in the right picture). The average score of 5 annotators was computed for each joke.}
    \label{fig:mean-scores}
\end{figure*}

Each joke has been rated by five different annotators. The scores were collected through \textbf{Telegram} bot by crowd sourcing. The annotators volunteered their time, dedicating a maximum of one hour. 
For each instance we calculated the mean score for five annotators, which is used in CleanComedy Gold datasets as human humor score, see Figure ~\ref{fig:mean-scores}. We also publish individual scores to facilitate further research on personalization of generative humor.

With the consent of the annotators we also publish their anonymized data, such as age, gender (female, male and other), education level (secondary vocational, bachelor's degree, specialist degree, master's degree, PhD) and the language fluency level for both languages (from A1 to C2 + native).

\section{Models}

In our experiments, we fine-tune a pre-trained version of a multilingual large language model \href{https://huggingface.co/meta-llama/Llama-3.1-8B}{meta-llama/Llama-3.1-8B}. We utilize the pre-train version of the model because we believe that in this setting the influence of our dataset on the final result increases, while using the instruction model \href{https://huggingface.co/meta-llama/Llama-3.1-8B-Instruct}{meta-llama/Llama-3.1-8B-Instruct} increases the dependence on both the model itself and the prompt selected for it.

Our prompt for fine-tuning models looks like this:
\begin{figure}[H]
    \begin{center}
    \mdfsetup{%
        middlelinecolor=black,
        middlelinewidth=1pt,
        backgroundcolor=gray!10,
        roundcorner=5pt,
        frametitlerule=true
        }
    \begin{mdframed}
    \texttt{<|begin\_of\_text|> \\
    <|reserved\_special\_token\_\{i\}|> \\
    \{text\}}
    \end{mdframed}
    \end{center}
\end{figure}

where \texttt{<|begin\_of\_text|>} is the start of text token, \texttt{<|reserved\_special\_token\_\{i\}|>} is the language token (i=0 for English and i=1 for Russian language), and \texttt{\{text\}} is a joke text.

For fine-tuning the model, we use LoRA \cite{hu2021lora}. Instead of updating all the weights in every layer, we only train low-rank approximations (with a rank of 4) for all layers except for the classification and embedding layers, which we train fully.

We fine-tune the model in two stages: \textbf{Supervised Fine-Tuning} and \textbf{LLM Alignment}. At the first stage, we train \href{https://huggingface.co/meta-llama/Llama-3.1-8B}{meta-llama/Llama-3.1-8B} to imitate the jokes of our large datasets. At the second stage, we additionally train the model obtained at the previous stage to generate funny examples from our small annotated dataset and not generate unfunny ones. At both stages, optimization is performed using the Adam algorithm, coupled with a cosine learning rate scheduler. The batch size per these two stages is equal to 64.

\subsection{Supervised Fine-Tuning}
\label{subsec:sft}

At this stage, we train \href{https://huggingface.co/meta-llama/Llama-3.1-8B}{meta-llama/Llama-3.1-8B} using the datasets described in \hyperref[subsec:data_processing]{Data processing} subsection.

The model's training employs the standard language modeling loss function. A fine-tuning phase takes 2 epochs using a maximum learning rate of 1e-05.

\subsection{LLM Alignment}
\label{subsec:alignment}

At this stage, we additionally train the model obtained at the previous stage using the CleanComedy Gold dataset.
We calculate the level of humor in a joke by averaging the annotators' ratings in this one. Since these average ratings are between 1 and 5, we apply a linear transformation to map the scores to a 0-to-1 scale to get soft labels for sequence classification task.

The model is trained using binary cross-entropy as the loss function for binary classification tasks inspired by \textbf{DPO} \cite{rafailov2024directpreferenceoptimizationlanguage} and \textbf{SimPO} \cite{meng2024simposimplepreferenceoptimization}. Our method differs from these methods by using soft positive/negative labels instead of requiring a dataset of chosen-rejected answer pairs:
$$t = (s - 1) / 4,$$
$$loss = -\sum_{(t, p) \in B}(t \log(p) + (1 - t) \log(1 - p)),$$
where $s$\footnote{The range of $s$ varies from 1 to 5, so $t$ falls between 0 and 1 and can be interpreted as the target probability for classification task using soft labels.} is an average humor score of a joke from the annotated dataset and $p$ is probability of the same joke according to the language model that is trained.

A fine-tuning phase takes 5 epochs using a maximum learning rate of 4e-07.

\section{Results}

For both languages, we sample 100 examples from each of the following six groups:

\begin{enumerate}[nosep]
    \item
    \textbf{LLaMA 3.1 8B (Supervised Fine-Tuned).}
    See \hyperref[subsec:sft]{Supervised Fine-Tuning} section for training details.
    \item
    \textbf{LLaMA 3.1 8B (Supervised Fine-Tuned + Aligned).}
    See \hyperref[subsec:alignment]{LLM Alignment} section for training details.
    \item
    \textbf{LLaMA 3.1 8B (Instruct).}
    We use following prompts for two languages.
    \begin{figure}[H]
    \begin{center}
    \mdfsetup{%
        middlelinecolor=black,
        middlelinewidth=1pt,
        backgroundcolor=gray!10,
        roundcorner=5pt,
        frametitlerule=true
        }
    \begin{mdframed}
    \textbf{System prompt:} \\
    \textit{You are a comedian with a lot of experience. Your income level and future career depend on the level of humor in jokes. Since your audience is educated, it's best to avoid using well-known jokes. \\
    } \\
    \textbf{User prompt:} \\
    \textit{Write a funny short joke for your performance. I only need the text of this joke and nothing else. 
    }
    \end{mdframed}
    \end{center}
\end{figure}
    \item
    \textbf{GPT-4o.}
    Since we don't have access to the GPT-4o API, we use the \textbf{default system prompt} and generate all 100 jokes of the class at one time, using the following prompt:
    \begin{figure}[H]
    \begin{center}
    \mdfsetup{%
        middlelinecolor=black,
        middlelinewidth=1pt,
        backgroundcolor=gray!10,
        roundcorner=5pt,
        frametitlerule=true
        }
    \begin{mdframed}
    \textbf{User prompt:} \\
    \textit{Hi! Imagine that you are a comedian with a lot of experience. Write 100 funny short jokes for your performance. Your income level and future career depend on them. 
    }
    
    \end{mdframed}
    \end{center}
\end{figure}
    \item
    \textbf{Unfiltered Dataset.}
    In order to understand how the levels of humor and toxicity in the dataset are changed after toxicity filters, we also take 100 samples from the dataset without toxicity filters.
    \item
    \textbf{Clean Dataset.}
    In order to understand how large language models are able to learn humor from humorous data, we also take 100 samples from the dataset as some reference for comparing humor scores.
    
\end{enumerate}

\begin{table*}[h!]
  \centering
  \begin{tabular}{ccc}
    Model     & Humor score \textuparrow    & Toxicity percentage  \textdownarrow  \\
    \midrule
    \multicolumn{3}{c}{English}                   \\
    \midrule
    LLaMA 3.1 8B (Supervised Fine-Tuned)  & $2.11 \pm 1.2$ & $13.38$ \\
    LLaMA 3.1 8B (Supervised Fine-Tuned + Aligned) & $2.02 \pm 1.08$ & $15.38$ \\
    LLaMA 3.1 8B (Instruct) & $2.65 \pm 1.23$ & $\textbf{3.97}$ \\
    GPT-4o & $\textbf{3.02} \pm 1.3$ & $9.27$ \\
    Unfiltered Dataset & $2.72 \pm 1.42$ & $26.09$ \\
    Clean Dataset & $2.96 \pm 1.36$ & $11.41$ \\
    \bottomrule
    \multicolumn{3}{c}{Russian}                   \\
    \midrule
    LLaMA 3.1 8B (Supervised Fine-Tuned) & $1.68 \pm 1.09$ & $4.01$ \\
    LLaMA 3.1 8B (Supervised Fine-Tuned + Aligned) & $1.74 \pm 1.06$ &  $\textbf{3.3}$ \\
    LLaMA 3.1 8B (Instruct) & $2.07 \pm 1.27$ & $4.26$ \\
    GPT-4o & $2.38 \pm 1.23$ & $4.67$ \\
    Unfiltered Dataset & $2.77 \pm 1.4$ & $20.93$ \\
    Clean Dataset & $\textbf{2.84} \pm 1.44$ & $11.37$ \\
    \bottomrule
  \end{tabular}
    \caption{Comparison of average metrics for different groups of jokes. Standard deviations are also calculated for the humor scores. Each group includes 100 examples, with each example rated by at least 3 people. The toxicity percentage ranges from 0 to 100, while the humor score ranges from 1 to 5.}
    \label{Comparison-of-Groups}
\end{table*}

We employ ancestral sampling with a temperature of 0.5 for all generated examples, except for the \textbf{LLaMA 3.1 8B (Instruct)} model for English, where we use a temperature of 0.9 to mitigate its tendency towards generating absolutely identical texts at lower temperatures. Additionally, for the \textbf{LLaMA 3.1 8B (Instruct)} generations, we remove semantic duplicates using the SBERT model in the same manner as during the dataset deduplication stage, applying the same cutoff threshold for identifying duplicates\footnote{We do this for two languages.}.

Each of the 600 sampled examples was evaluated by 3 individuals, following a process similar to that described in \hyperref[subsec:dataset_annotation]{Dataset Annotation} subsection. In this case, in addition to rating each joke, the evaluators were also asked to assess its level of toxicity or inappropriateness\footnote{We ask: "Do you find this text vulnerable or inappropriate?" An annotator can answer "yes" or "no".}. The average ratings for each group can be found in Table \ref{Comparison-of-Groups}. For both English and Russian, the toxicity percentage in our clean datasets is half that of the originally collected data, demonstrating the effectiveness of the aforementioned filtering process. The data on the demographics of the annotators can be found in Figures \ref{fig:ages}, \ref{fig:genders}, \ref{fig:education-level} and \ref{fig:language-level}. 

The evaluation of humor generation models reveals interesting trends across different setups for both English and Russian datasets. 

For English, among the models, GPT-4o achieves the highest humor score, closely followed by the Clean Dataset of human-created jokes. The LLaMA 3.1 8B (Instruct) model also performs well, with a humor score, but significantly lags behind GPT-4o. Models fine-tuned on the CleanComedy dataset, such as Supervised Fine-Tuned and Aligned, show lower humor scores, reflecting the challenges in adapting LLMs to humor tasks using limited or filtered data. In terms of toxicity, the LLaMA 3.1 8B (Instruct) achieves the lowest percentage, significantly outperforming both the unfiltered dataset and the clean dataset.

Similar trends are observed in Russian, where Clean Dataset achieves the highest humor score, followed by GPT-4o. The LLaMA 3.1 8B (Instruct) model scores slightly lower than its English counterpart, suggesting room for improvement in handling humor in Russian. Fine-tuned models for Russian also show lower humor scores. The Aligned model, however, demonstrates the lowest toxicity percentage, indicating that filtering and alignment processes are particularly effective in producing clean content for Russian.

The age histograms in Figure \ref{fig:ages} show the majority of our annotators are aged 20-30, which skews the humor perception toward this age group. It is also important to point out that there was almost no native speakers of English among our volunteers (Figure \ref{fig:language-level}), which has significantly influenced the humor scores obtained for the English jokes. We theorize that the English jokes generated by our models were rated higher than Russian due to this linguistic bias. The humor scores for the Russian generative jokes are lower for the models trained on the filtered data because they could have been deemed more boring for the annotators, as the humor scores are lower for the jokes with lower toxicity percentage.

\section{Conclusion}
In this research, we share the insights we gained while experimenting with prompting and fine-tuning models for humor generation.

This paper presents CleanComedy, a novel and ethically curated dataset for humor generation and evaluation in English and Russian. Through rigorous data filtering and annotation processes, we address the common challenges in datasets, such as toxicity and redundancy. By fine-tuning a LLM with this dataset, we demonstrated its capability to generate contextually appropriate and ethically aligned humor. 

The results provided by the annotators of our experiments underscore the importance of alignment techniques in improving the quality and relevance of generated humor. While our models perform well within the scope of the dataset, challenges remain in generating humor that adapts different languages and cultures.

This study contributes to the field by introducing methodologies that prioritize ethical considerations while advancing computational humor research. However, humor's subjective and cultural nature requires ongoing attention to ensure AI-generated content remains inclusive, engaging, and responsible.

\section*{Limitations}

The English dataset faces challenges related to the quality and diversity of humor instances. While significant efforts are made to remove offensive and inappropriate content from the datasets, the filtering processes may not be ideal. Enhanced filtering techniques and ongoing monitoring are essential to mitigate this risk. Moreover, future research should focus on developing more advanced generative techniques to enhance the coherence and creativity of the outputs.

When training language models, we have only experimented with PEFT methods. Judging by the quality of the resulted generations, the models would benefit greatly from full fine-tuning. Another limitation of this study is the size of the dataset which comprised only  \(\sim \)40k samples for each language. It seems obvious that collecting a bigger and higher quality humor dataset could further advance humor generation. For example, transcribing popular humor shows seems to be an interesting idea for humor data collection.

Humor is inherently connected to cultural and contextual details, which can be challenging for language models to understand and generate. The models we trained for English and Russian show different types of jokes, thus, they might not perform well across other cultural contexts and languages, rising the need for culturally adaptive humor generation techniques.

\section*{Ethical Statement}
The ethical risks that the computational humor research carries are discussed in this section ensure responsible development of AI technologies.

Computational humor systems may mimic biases present in training data. Despite multiple stage filtering process and mitigation strategies to prevent the spread of stereotypes and discriminatory content, generated humorous texts still can sometimes result in offensive or harmful content.

There is a risk that computational humor systems could be misused for malicious purposes, such as cyberbullying, harassment, or spreading misinformation, thus, we claim that our systems are not intended for ant malicious applications.

The rise of computational humor may impact on human writing and entertainment. It is essential to consider societal impacts and strive for a balance that supports human creativity while leveraging technological advancements.

 \section*{Acknowledgments}

We would like to express our gratitude to the volunteer crowdworkers who dedicated their time and effort to evaluate our generative models.

\bibliography{custom}

\appendix
\clearpage
\onecolumn
\section{Appendix}
\label{sec:appendix}

\begin{table*}[h]
    
    \centering
    \begin{tabular}{p{8cm}p{4cm}}
        \hline
        Text & Drawback \\
        \hline
        well.... [https://imgur.com/gallery/2CmdahS] (https://imgur.com/gallery/2CmdahS) & links \\
        \begin{CJK*}{UTF8}{gbsn} 占占占人 占占占人 占占点 占点占 点占占 …占占占 \end{CJK*} & not latin characters \\
        A veteran walks into a bar. 12 people in the bar [removed] & [removed] \\
        Chinas president is so bad beacuse.... [Deleted] & [deleted] \\
        lock lock me, i dare you. & not a joke (excessively short) \\
        Test post Test & not a joke (excessively short) \\
        meat meat meat meat meat meat meat meat meat meat meat meat meat meat meat meat meat meat meat meat meat meat meat meat meat meat meat meat meat meat meat meat meat meat meat meat meat meat meat meat meat meat meat meat meat meat meat meat meat meat meat meat meat meat meat meat meat meat meat meat meat meat meat meat meat meat meat meat meat meat meat meat meat & repetition (excessively long) \\
        Mars is Earth's second moon Mars is Earth's second moon Mars is Earth's second moon Mars is Earth's second moon Mars is Earth's second moon... (and so on) & repetition (excessively long) \\
    \hline
    \end{tabular}
    \caption{Different types of noise before initial preprocessing of the English data. All the provided examples were removed after initial preprocessing that includes the removal of corrupted characters and excessively short or long examples (shorter than 30 or longer than 150 characters).}
    \label{en-rubbish}
\end{table*}

\begin{table*}[h]
  
  \centering
  \begin{tabular}{p{8cm}p{2cm}p{2cm}}
    \hline
    Joke   &  ruBERT Conversatinal Toxicity Classifier & Multilingual Detoxify  \\
    \hline
    Если бы я была принцессой, то моим замком был бы ликеро-водочный завод. \textit{(If I were a princess, my castle would be a distillery.)} & 0 & 0.01  \\
    Где твоя грудь? Ты ее спугнул. \textit{(Where are your breasts? You scared them away.)} & 1 & 0.83  \\
    Мы не будем больше встречаться... Что, тараканы в твоей голове проголосовали против меня? \textit{(We won't be dating anymore... What, the bugs in your head voted against me?)} & 0 & 0.73  \\
    
    \hline
  \end{tabular}
  \caption{Different toxicity classifiers results for CleanComedy Russian, 1 -- for toxic, 0 -- neutral, for ruBERTConv we considered jokes with score more than 0.1 to be inappropriate.}
  \label{rus-detox}
\end{table*}


\begin{table*}[h]
  
  \label{duplicates}
  \centering
  \begin{tabular}{p{12cm}l}
    \hline
    Duplicate    & Cosine   \\
    \hline
    \multicolumn{2}{c}{English}                   \\
    \hline
    \multicolumn{2}{c}{what's a rock group with four guys that don't sing? mount rushmore}                   \\
    \hline
    What rock group has four guys who can't sing?  Mount Rushmore. & 0.922  \\
    who's an all male rock group that doesn't sing? mount rushmore .    & 0.8857    \\
    what do you call a rock group with no bassist, drummer, singer or guitarist? Mount Rushmore & 0.741 \\
    Q: What do you call a male quartet? A: Three men and a tenor.   & 0.528 \\

    \hline
    \multicolumn{2}{c}{Russian}                   \\
    \hline
    \multicolumn{2}{c}{а зачем в низу эскалатора бабулька в будке сидит? - она там педали крутит.}   \\
    \multicolumn{2}{c}{\textit{(Why is grandma sitting in a booth at the bottom of the escalator? - She's pedalling there.)}}                  \\
    \hline
    А знаете, зачем в метро внизу эскалатора бабулька в будке? Она там педали крутит. \textit{(Do you know why there is a granny in a booth in the subway at the bottom of the escalator? She's pedalling there.)} & 0.928 \\
    \hline
    \multicolumn{2}{c}{Привет, как дела? Вот все тебе надо знать.}                   \\
    \multicolumn{2}{c}{\textit{(Hi, how are you? You gotta know everything, don't you?)}}     \\
    \hline
    - Привет! Как дела? - Порадовать тебя нечем. - Как? Неужели всё хорошо? \textit{(Hello! How are you? - I can't please you with anything. - How come? Is everything really good?)}  & 0.9033    \\
    
    \hline
  \end{tabular}
  \caption{Cosine similarity between duplicates}
\end{table*}

\clearpage

\begin{table*}[h]
  
  \label{deberta}
  \centering
  \begin{tabular}{p{11cm}p{2cm}}
    \hline
    Joke    & Label   \\
    \hline
    \multicolumn{2}{c}{English}                   \\
    \hline
    Romeo: Your cheeks are like petals. Juliet: Really? Romeo: Yes, bicycle pedals.   & neutral  \\
    What would one call a movie about meth addictions? Need for speed.  & drugs \\
    what do you call ghosts that haunt liquor stores? spirits   & alcohol   \\
    What do Syrian refugees eat for breakfast? Syrial!   & racist    \\
    Bought my epileptic girlfriend a strobe light for her birthday. She will have a fit when she sees it.    & offending \\
    Why can't you have Christmas dinner in the EU? Because there is no turkey.    & politics  \\
    \hline
    \multicolumn{2}{c}{Russian}                   \\
    \hline
    Иногда я выпиваю стакан воды, просто для того что бы удивить свою печень. \textit{(Sometimes I drink a glass of water, just to surprise my liver.)} & alcohol \\
    - Знаешь, я целый месяц ходила на фитнес!... - И сколько сбросила? - 12 тысяч!!!... \textit{(- You know, I went to fitness for a whole month!... - And how much did you lose? - 12 hundred!!!...)} & neutral   \\
    Лучше было бы, если бы народные избранники отвечали не за свои предвыборные слова, а за свои послевыборные дела. \textit{(It would be better if the people's representatives were responsible not for their pre-election words, but for their post-election deeds.)}   & politics  \\
    99 процентов моей подготовки к экзамену это курение с грустным лицом. \textit{(99 of my exam preparation involves smoking with a sad face.)}    & drugs \\
    Бомжи вышли на демонстрацию против новых технологий. По их словам, в коробке из-под плоского телевизора вообще жить невозможно. \textit{(Homeless people demonstrated against new technologies. According to them, it is generally impossible to live in a flat-screen TV box.)}    & offending \\
    Американский турист ходит с гидом по Лондону. — Все тут у вас такое маленькое, зажатое, — говорит он. — Это здание, например, было бы в Америке раз в десять выше. — О, конечно, сэр! Это-же психиатрическая клиника. \textit{(An American tourist walks with a guide around London. “Everything here is so small and cramped,” he says. “This building, for example, would be ten times taller in America.” - Oh, of course, sir! This is a psychiatric clinic.)}  & racist    \\
    
    \hline
  \end{tabular}
  \caption{Sentiment analysis}
\end{table*}


\begin{table*}[h!]
  
  \label{clusters}
  \centering
  \begin{tabular}{p{8cm}p{5cm}}
    \hline
    Joke    & Cluster name   \\
    \hline
    \multicolumn{2}{c}{English}                   \\
    \hline
    Idea:A Transformers movie that can transform into a much better movie.    & movie   \\
    'Why did voldermort used Twitter instead of Facebook? Because he only had followers. Not friends.    & facebook  \\
    What did the Jewish man say to himself on a hot day? I should be used to being in an oven by now.    & jews \\
    What kind of bee can't be understood? A mumble bee ! & bee   \\
    what do you call a dead bee?a was   & bee   \\
    What happens to someone who gets attacked by bees? They get bee'd up & bee   \\
    I just tried coffee for the first time... To be honest, it wasn't my cup of tea...  & coffee \\
    What do you call a cat in love? Romeow   & cat  \\
    \hline
    \multicolumn{2}{c}{Russian}                   \\
    \hline
    Вся жизнь — футбол, а ты в ней — сборная России. \textit{(All life is football, and in it you are the Russian national team.)}  & футбол \textit{(football)}    \\
    Доктор, помогите мне. У меня проблема. Я часто ошибаюсь в людях. Я не доктор. \textit{(Doctor, help me. I have a problem. I often make mistakes in people. I'm not a doctor.)}  & доктор \textit{(doctor)}    \\
    Поймал мужик золотую рыбку хочу машину и завод. Рыбка: хорошо но в кредит или по лизингу? Старик: так выбирай на подсолнечном или сливочном? \textit{(A man caught a goldfish and tells it that he wants a car and a factory. Goldfish: okay, do you want to loan or to lease it? Old man: so choose oil or fat?)}  & рыба \textit{(fish)}    \\
    Я не понимаю фразу: Устал, как собака. Моя собака спит, ест и гуляет. Я бы сам от такой жизни не отказался. \textit{(I don't understand the phrase: Tired like a dog. My dog sleeps, eats and walks. I myself would not refuse such a life.)} & собака \textit{(dog)}   \\
    В компании со мной обычно смеются не над шуткой, а над тем, как я смеюсь над шуткой. \textit{(People with me usually laugh not at the joke, but at the way I laugh at the joke.)}   & юмор \textit{(humour)}    \\
    \hline
  \end{tabular}
  \caption{Sentiment analysis}
\end{table*}

\clearpage

\begin{figure*}[t]
    \centering
    \includegraphics[width=0.8\textwidth]{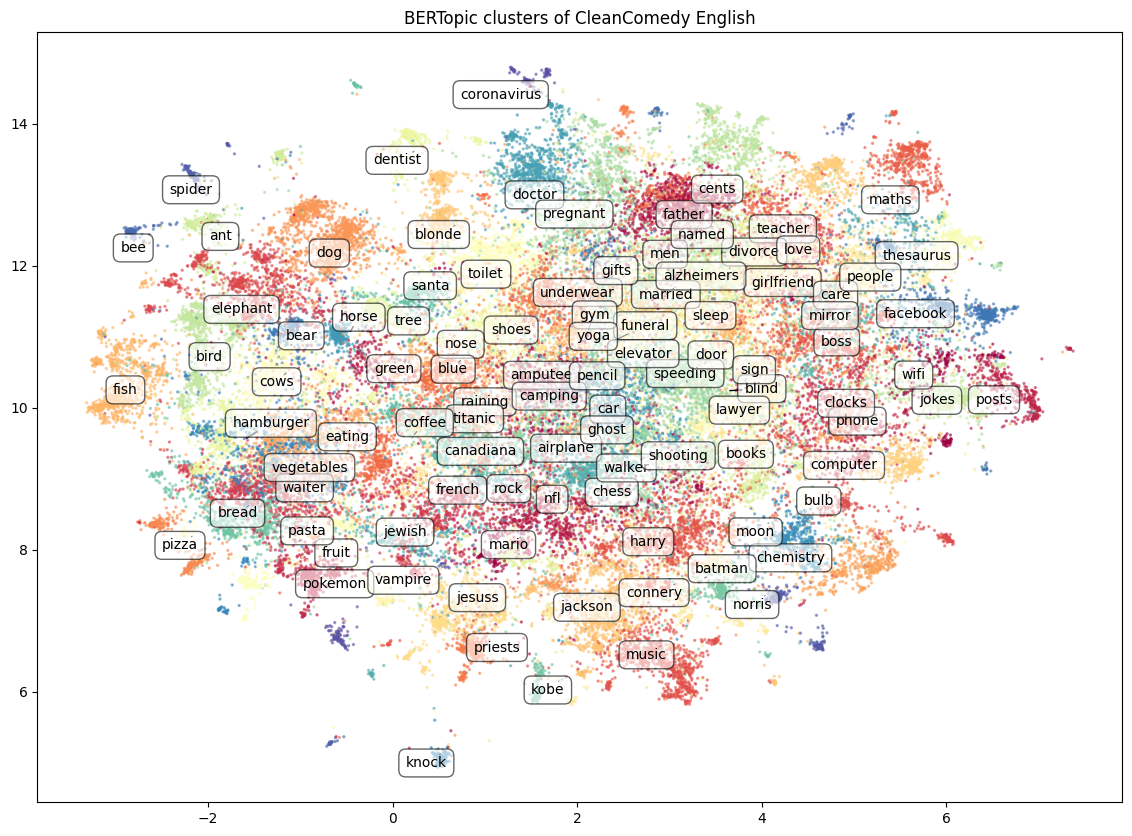}
    \caption{Topic modeling for CleanComedy English.}
    \label{fig:clusters-en}
\end{figure*}

\begin{figure*}[b]
    \centering
    \includegraphics[width=0.8\textwidth]{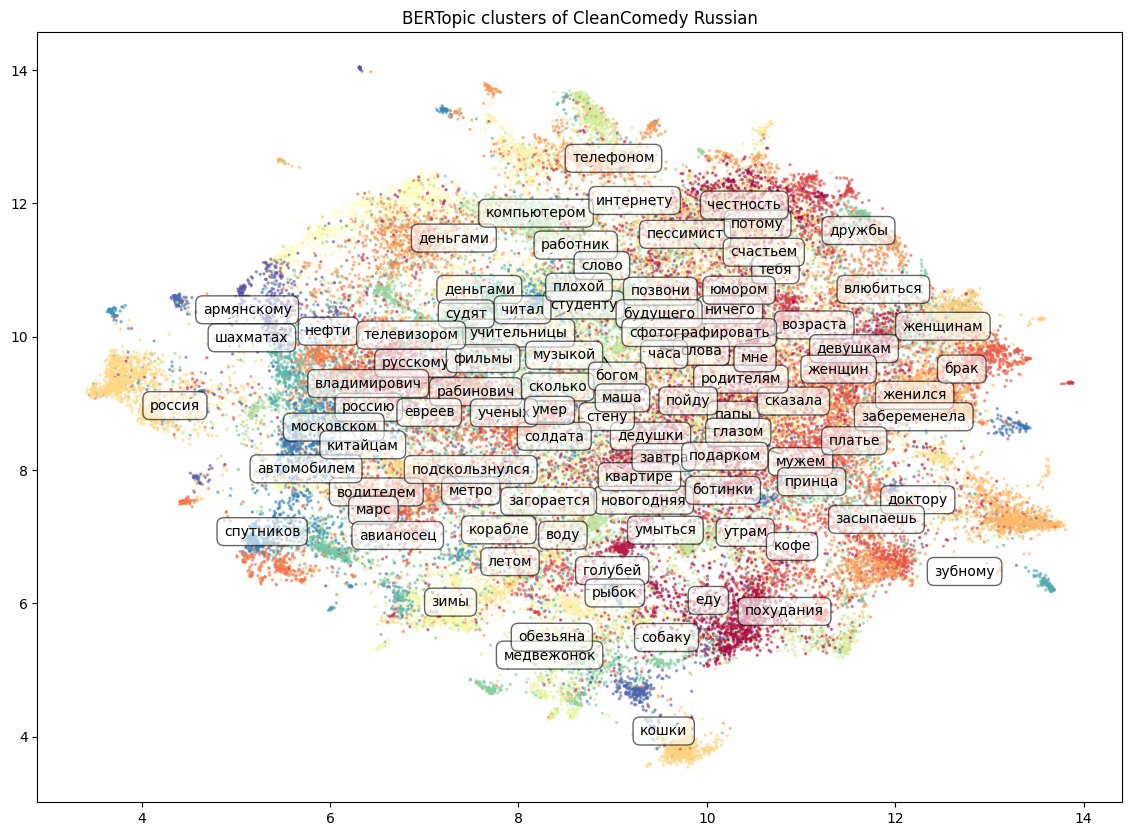}
    \caption{Topic modeling for CleanComedy English.}
    \label{fig:clusters-ru}
\end{figure*}

\clearpage

\begin{figure*}[h!]
    \includegraphics[width=0.5\textwidth]{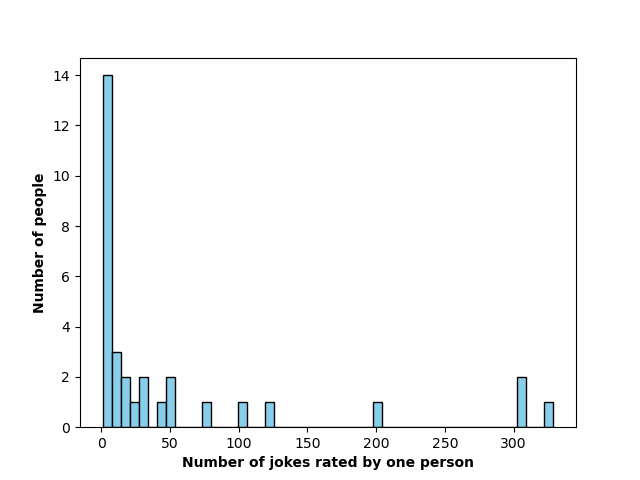}\hfill
    \includegraphics[width=0.5\textwidth]{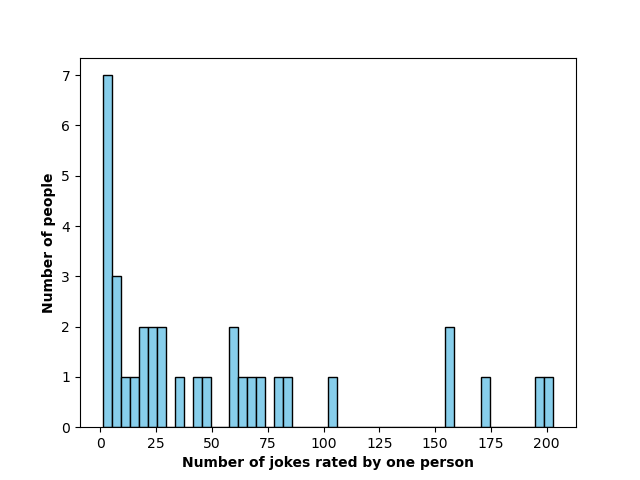}\hfill
    \caption{Number of jokes rated by one person for English (in the left picture) and Russian (in the right picture). Only annotators with at least one score are taken into account.}
    \label{fig:rated-by-one-person}
\end{figure*}

\begin{figure*}[h!]
    \includegraphics[width=0.5\textwidth]{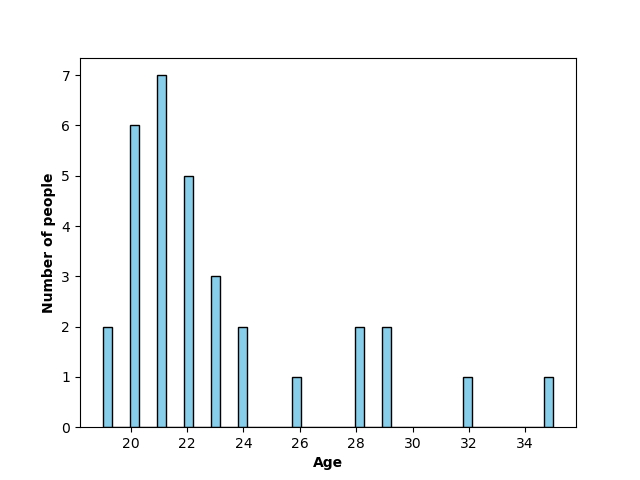}\hfill
    \includegraphics[width=0.5\textwidth]{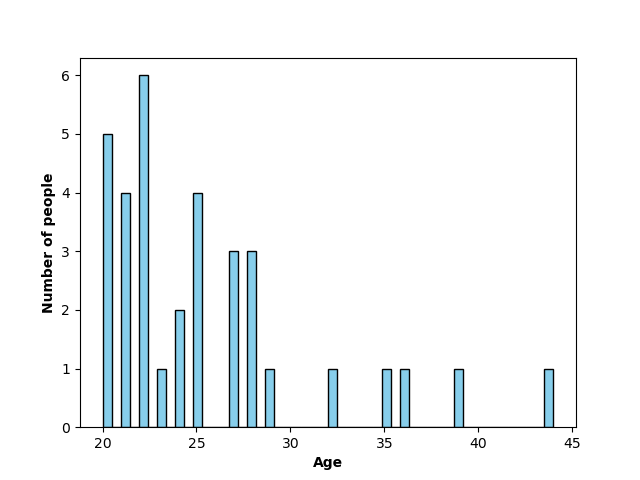}\hfill
    \caption{Age of annotators for English (in the left picture) and Russian (in the right picture). Only annotators with at least one score are taken into account.}
    \label{fig:ages}
\end{figure*}

\begin{figure*}[h!]
    \includegraphics[width=0.5\textwidth]{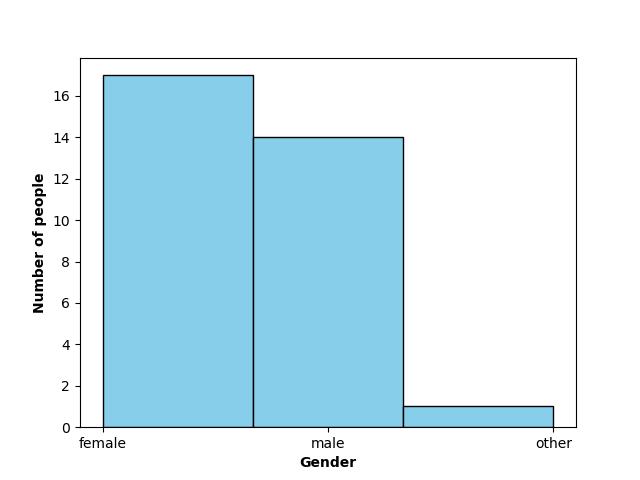}\hfill
    \includegraphics[width=0.5\textwidth]{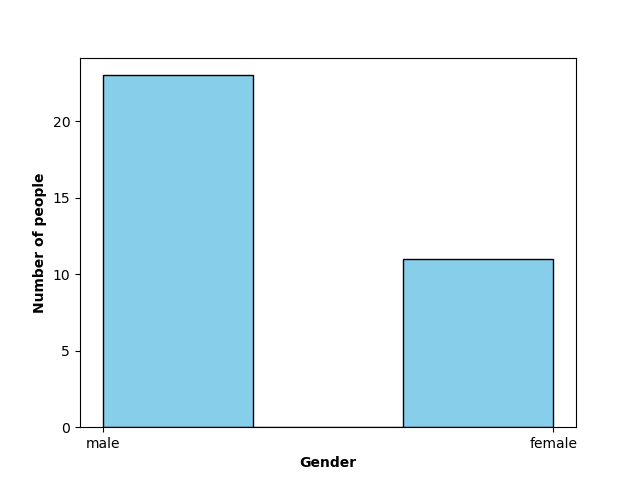}\hfill
    \caption{Gender of annotators for English (in the left picture) and Russian (in the right picture). Only annotators with at least one score are taken into account.}
    \label{fig:genders}
\end{figure*}

\begin{figure*}[h!]
    \includegraphics[width=0.5\textwidth]{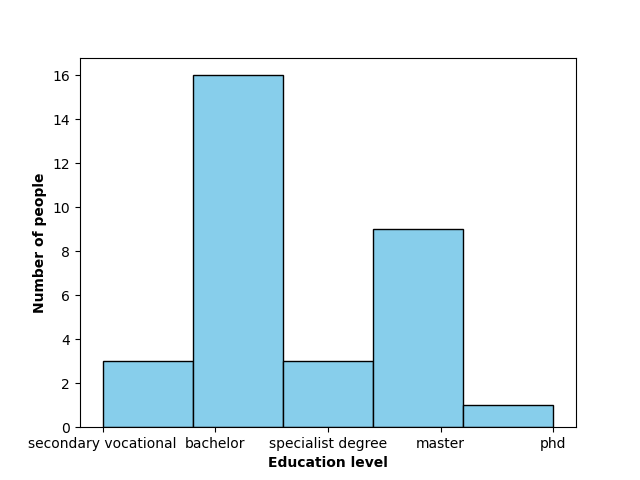}\hfill
    \includegraphics[width=0.5\textwidth]{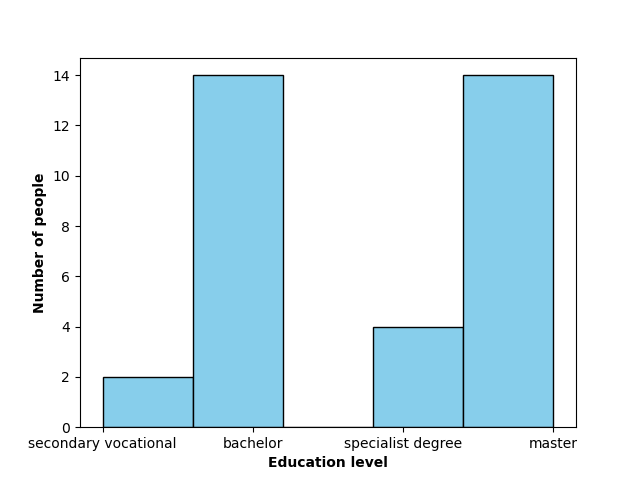}\hfill
    \caption{Education level of annotators for English (in the left picture) and Russian (in the right picture). Only annotators with at least one score are taken into account.}
    \label{fig:education-level}
\end{figure*}

\begin{figure*}[h!]
    \includegraphics[width=0.5\textwidth]{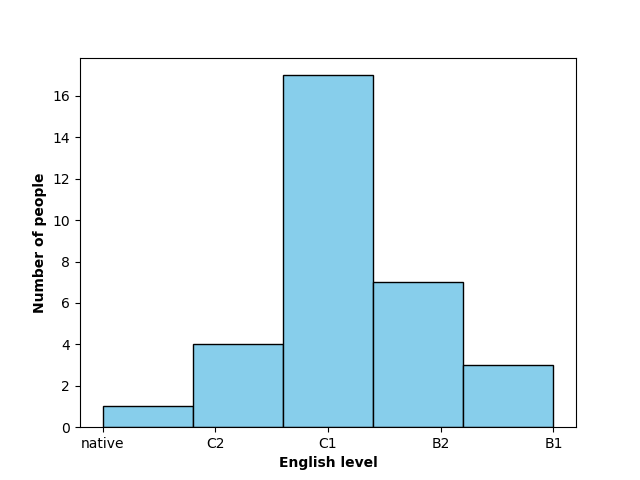}\hfill
    \includegraphics[width=0.5\textwidth]{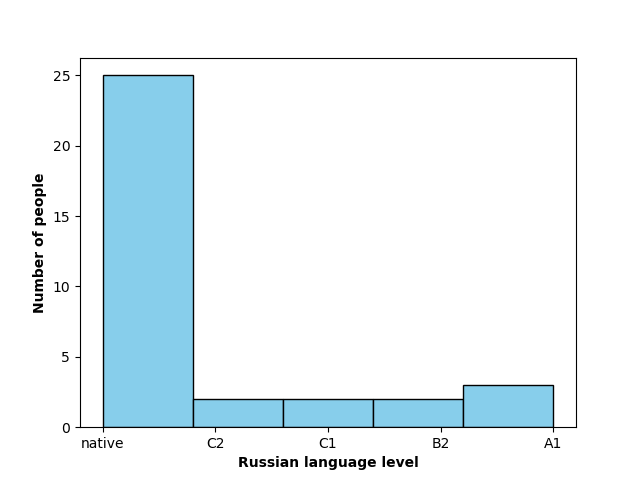}\hfill
    \caption{Language level of annotators. Only annotators with at least one evaluation point are taken into account.}
    \label{fig:language-level}
\end{figure*}

\end{document}